\documentclass{i4c-en}
\usepackage{algorithm}
\usepackage{algpseudocode}
\usepackage{graphicx}
\usepackage{booktabs}
\usepackage{multirow}
\usepackage{amssymb}
\usepackage{hyperref}
\usepackage{url}

\usepackage[switch]{lineno}
\usepackage{multirow,multicol}
\usepackage{makecell}
\usepackage{colortbl}

\begin{document}
	\thispagestyle{plain}	
	\begin{center}
		{\Large\bf The Maskability Index: Predicting Task--Objective Alignment in Pretrained Language Models}
		\vspace*{0.5cm}
		
			{\small Ahmad Pouramini}\index{Pouramini, Ahmad}\footnote{speaker}, 
	{\small Mahsa Afsharizadeh}\index{Afsharizadeh, Mahsa}\\
	{\small Department of Computer Engineering, Sirjan University of Technology}\\[2mm]
		
	\end{center}
	
	\vspace*{0.5cm}
	
	\hspace{-1cm}\rule{\textwidth}{0.2mm}
	
	\begin{abstract}
	Large-scale pretrained language models such as T5 and BERT have shown the ability to generate structured  knowledge, but their effectiveness depends strongly on how the prompting style aligns with the model’s pretraining objectives. We introduce the \textbf{Maskability Index (MI)}, a quantitative measure that estimates how well a knowledge relation benefits from masked-style prompting compared to prefix-style prompting in few-shot generative settings. MI is computed from DepthRank differences between masked and unmasked templates, offering a principled way to capture objective–template alignment. We validate MI on a wide range of relations from the ATOMIC2020 knowledge base completion benchmark and show that it correlates with downstream generation quality. Our results suggest that MI can guide the choice of template families and adaptation strategies when mining relational knowledge from pretrained models, particularly in low-resource conditions.
\end{abstract}

	\subject{68T50, 68Q32}
	
\hspace{-1cm}\rule{\textwidth}{0.2mm}

\section{Introduction}
Pretrained language models (PLMs) such as BERT \cite{devlin2019bert} and T5 \cite{raffel2020exploring} have been shown to encode large amounts of factual and commonsense knowledge, which can be surfaced through carefully designed prompts. Early studies, most notably LAMA \cite{petroni2019language}, demonstrated that masked cloze-style templates could probe relational knowledge directly from PLMs without additional fine-tuning. This line of work sparked extensive research on prompt-based evaluation and adaptation, revealing both the potential of PLMs as implicit knowledge bases and the striking sensitivity of performance to prompt wording and format \cite{jiang2020can,shin2020autoprompt,gao2021making}.

Subsequent work highlighted that this sensitivity is not accidental but stems from a deeper interaction between a model’s pretraining objective and the style of prompt used at inference time \cite{ul2, pouramini2024matching, brown2020language, liu2021pretrain}. Models trained with masked or denoising objectives (e.g., BERT, T5’s span corruption) are naturally aligned with cloze-style templates, while autoregressive models (e.g., GPT) favor left-to-right conditional generation \cite{raffel2020exploring,radford2019language}. Even within a single model such as T5—which combines a denoising objective with additional left-to-right training—different relations or tasks may be more recoverable under one template family than the other \cite{pouramini2024matching, ul2}. 

Recent advances in unified pretraining frameworks such as UL2 \cite{ul2} have shown that denoising, causal, and infilling objectives can be treated as part of a continuous design space rather than distinct paradigms. This perspective motivates our analysis of how different objectives align with downstream task templates through information-theoretic measures.

Zhang et al.\  introduced the notion of \emph{deep commonsense knowledge}, observing that many relational triples diverge significantly from natural language expression and thereby pose a challenge for language models. They operationalize this via DepthRank and perplexity metrics on triple-to-sentence conversions, and show that triples of higher depth (i.e., more “unnatural” ones) lead to significantly poorer model performance under naive prompting or classification setups \cite{zhang2021alleviating}. Building on this insight, we hypothesize that template–objective alignment (masked vs prefix prompting) will interact strongly with this depth: masked-style prompting may succeed for shallow, lexical relations, but fail for deep, inferential ones.  


To capture this interaction, we propose the \emph{Maskability Index (MI)}, a statistic that operationalizes template–objective alignment. MI is defined as the difference in DepthRank values obtained under masked versus prefix-generation templates, and it provides a direct estimate of whether a relation is more “mask-friendly” or “generation-friendly.” We evaluate MI on the ATOMIC2020 commonsense knowledge base \cite{atomic2020}, focusing on knowledge completion as a generative task. Our experiments show that MI correlates with downstream generation quality across relations and can guide the choice of template families and adaptation strategies, particularly in few-shot or low-resource settings.

Our contributions are as follows:
\begin{itemize}
	\item We highlight the role of pretraining–prompt alignment in knowledge base completion and frame it as a measurable phenomenon at the relation level.
	\item We propose the Maskability Index (MI), a novel statistic derived from DepthRank differences, which predicts whether masked-style or prefix-style prompting will be more effective.
	\item We validate MI on ATOMIC2020 relations, showing its ability to correlate with and even anticipate downstream performance, offering practical guidance for template selection and adaptation in generative knowledge completion.
\end{itemize}

\section{Empirical Motivation: DepthRank Variation Across Templates}

Pretrained language models (PLMs) can generate plausible tails given a head–relation pair, but the ease of generation varies widely across relations and prompting styles. Consider the task of \textbf{head--relation--tail generation}: given a head event or entity \(h\) and a relation \(r\), the model predicts one or more plausible tails \(t\) that complete the relation.  

To capture how accessible gold tails are in the model’s predictions, we use the concept of \textbf{DepthRank (DR)}. Intuitively, DR measures the rank of the gold tail tokens within the model’s predicted probability distribution. Lower DR indicates that a gold tail is near the top of the predicted list (easier to generate), while higher DR indicates it is more “buried” (harder to predict). DepthRank is formally defined in Section~\ref{sec:dr}, but here we illustrate its behavior with examples.

\begin{table}[h!]
	\centering
	\small
	\setlength{\tabcolsep}{5pt}
	\renewcommand{\arraystretch}{1.2}
	\begin{tabular}{llllc}
		\toprule
		\multicolumn{5}{c}{\textbf{Head:} PersonX works as a waitress } \\ 
		\midrule
		\textbf{Relation} & \textbf{Prediction} & \textbf{Target (tail)} & \textbf{Token Idx} & \textbf{DR} \\ 
		\hline
		\multirow{2}{*}{\makecell[c]{xIntent \\ (because she intends)}} 
		& \multirow{2}{*}{to be a waitress} 
		& to earn money & $1,53,1$ & $18.33$ \\ \cmidrule{3-5}
		&  & to make money & $1,7,1$ & $3$ \\ 
		\midrule
		
		\multirow{3}{*}{ \makecell[c]{xAttr \\ (he is seen as)}} 
		& \multirow{3}{*}{caring} 
		& exhausted & $1672$ & $1672$ \\ \cmidrule{3-5}
		&  & tired & $1136$ & $1136$ \\ \cmidrule{3-5}
		&  & hard working & $33,4$ & $17$ \\ 
		\bottomrule
	\end{tabular}
	
	\caption{
		Illustrative \texttt{T5-base} generations on ATOMIC2020 relations for the head \textit{“PersonX works as a waitress”}. 
		Each block shows a relation–prediction pair followed by plausible gold completions with their token indices and DepthRank (DR). 
		Lower DR indicates that the gold is nearer to the model’s top predictions, reflecting stronger internal alignment.
	}
	\label{table:examples}
\end{table}

Table~\ref{table:examples} presents a small set of \texttt{T5-base} generations under different ATOMIC2020 \cite{atomic2020} relations. For each head–relation pair, we show the model’s predicted continuation, the corresponding gold targets (reference tails), token indices of those gold targets in the model’s probability list, and the resulting DepthRank (DR) values, which represent the average rank position of the gold tail tokens in the prediction distribution. 

In each case, the relation is incorporated into the input template through a short natural phrase that verbalizes the semantic role of the relation (e.g., \textit{xIntent} $\rightarrow$ “because she intends,” \textit{xAttr} $\rightarrow$ “she is seen as”). These verbalizers help transform abstract relational labels into fluent, interpretable textual contexts compatible with the models pre-training objectives. The complete mapping between relations and their corresponding natural-language phrases is listed in Appendix~Table~\ref{table:tuples}.

Importantly, DepthRank is calculated over \emph{gold target tokens}, not over the model’s own prediction. A prediction always has rank zero by definition because it corresponds to the highest probability token chosen by the model at each step. In contrast, DepthRank reveals how far the plausible gold targets are from the model’s top predictions. A lower DR indicates that the gold tokens appear closer to the model’s top predictions, reflecting stronger internal alignment and better relational consistency.

Relations like \textit{xIntent} typically involve more compositional structures (e.g., infinitival clauses beginning with “to”), yet these yield lower DepthRank values. This pattern arises because the model strongly associates \textit{xIntent} with the token “to,” constraining subsequent predictions to a narrow verb space with relatively low rank positions. Conversely, adjective completions for \textit{xAttr} are more lexically diverse and less predictable, leading to greater DepthRank variability.

The relations presented in Table~\ref{table:examples} were computed using the \textit{Prompting} method, in which the decoder generates a continuation of the head–relation pair provided to the encoder. This setting aligns with the language modeling (LM) objective used during pretraining, where the model learns to predict the next token in a sequence.
An alternative setting is \textit{MaskedPrompting}, where a mask token is inserted in place of the missing part, and the decoder generates the corresponding word or phrase. This formulation aligns with the denoising pretraining objective used in encoder–decoder architectures such as T5, where the model learns to reconstruct masked spans of text.

\begin{figure}[h!]
	\centering
	\includegraphics[width=0.9\linewidth]{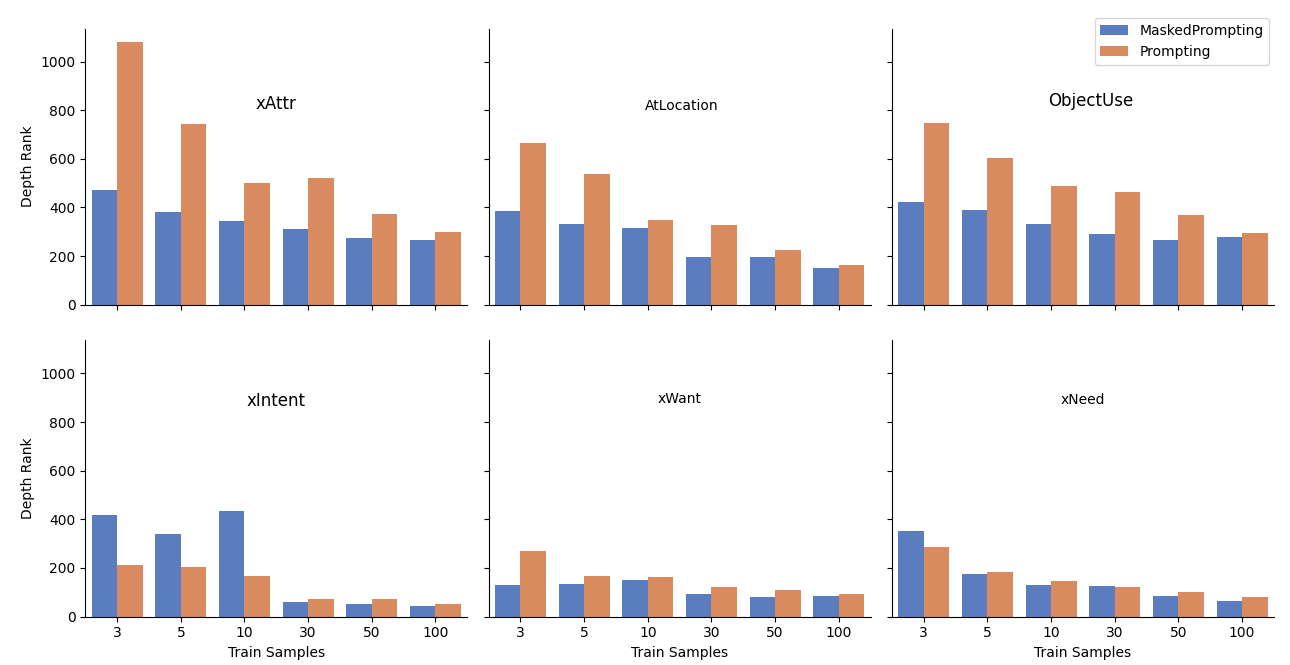}
	\caption{Average DepthRank of different ATOMIC2020 relations as a function of the number of training samples, comparing \textit{Prompting} and \textit{MaskedPrompting} templates. Mask-style templates often reduce DR for lexical relations in low-resource settings, motivating the need for a metric to quantify template--objective alignment.}
	\label{fig:deep}
\end{figure}

Figure~\ref{fig:deep} presents the average DR for multiple head--relation instances under two template families: \textit{Prompting} (prefix-style) and \textit{MaskedPrompting} (mask-style). The sample size was varied from 3 to 100. Some relations, like \textit{AtLocation} and \textit{xAttr}, achieve lower DR under masked prompting, particularly in few-shot scenarios. Conversely, relations such as \textit{xIntent} and \textit{xNeed} show smaller differences, or even slightly favor prefix prompting.  

These observations highlight a systematic phenomenon:  

\begin{itemize}
	\item Relations with short lexical tails (e.g., adjectives or common object uses) benefit from masked prompting, achieving lower DR.  
	\item Relations requiring longer, compositional, or inferential tails (e.g., xIntent, xNeed) are less amenable to single-span mask recovery, showing smaller DR gains under masked templates.  
\end{itemize}

This motivates the need for a relation-level metric that quantifies how well a relation aligns with a template family or pretraining objective. In the following Methodology section, we formally define DepthRank and introduce the \textbf{Maskability Index (MI)}, which leverages these differences to predict the most effective prompting style for a relation.

\section{Methodology} \label{sec:method}

\subsection{Task Definition}

We frame our study as a general knowledge base completion (KBC) problem: given an incomplete relational triple consisting of a head entity or event \( h \) and a relation \( r \), the model must generate a plausible tail \( t \) that completes the relation. 

Formally, each instance is represented as a tuple \((h, r, t)\), where \(h\) and \(t\) are natural language expressions and \(r\) denotes a semantic relation describing causal, attributive, or situational knowledge (e.g., \textit{is used for}, \textit{wants}, \textit{needs}). 

\subsection{DepthRank} \label{sec:dr}
Following \cite{zhang2021alleviating}, we measure how highly a pretrained model ranks the gold tail tokens for a triple presented under a given template family.

Let a triple be tokenized as

\[
S = \{h_{1:m},\; r_{1:n},\; t_{1:k}\},
\]
where \(h\) is the head, \(r\) the relation (or relation phrase), and \(t\) the tail (target sequence), with token counts \(m,n,k\). For each tail token \(t_i\) we compute the rank (index) of the correct token within the model's sorted probability list:
\[
\mathrm{Index}\bigl(t_i \mid h_{1:m}, r_{1:n}, t_{<i}\bigr).
\]
The DepthRank of the full target tail sequence is
\[
\mathrm{DepthRank}(S) = \frac{1}{k}\sum_{i=1}^{k}\mathrm{Index}\bigl(t_i \mid h_{1:m}, r_{1:n}, t_{<i}\bigr).
\]
Lower DepthRank means the gold tokens are nearer the top of the list (easier to predict).

	\subsection{Maskability Index (MI)}
	DepthRank can be computed under two template families:
	
	\begin{itemize}
		\item \textbf{Prompting (P)}: unmasked prefix templates that rely on conditional generation (e.g., ``PersonX cooks, before that they need \(\dots\)'').
		\item \textbf{MaskedPrompting (MP)}: templates that place an explicit unique mask token(s) corresponding to the tail and expect the model / decoder to recover the masked span (aligned with denoising objectives).
	\end{itemize}
	
	For a relation \(r\) and a chosen few-shot sample size \(n\), define the mean DepthRank across an \(n\)-sample under each template family:
	\[
	DR_{\mathrm{Prompting}}(r,n), \qquad DR_{\mathrm{MaskedPrompting}}(r,n).
	\]
	Then the Maskability Index is defined as the relative DepthRank improvement of masked prompting over prompting:
	\begin{equation}\label{eq:mi}
		\mathrm{MI}(r,n) \;=\; \frac{DR_{\mathrm{Prompting}}(r,n) \;-\; DR_{\mathrm{MaskedPrompting}}(r,n)}{DR_{\mathrm{Prompting}}(r,n)}.
	\end{equation}
	
	Interpretation:
	\begin{itemize}
		\item \(\mathrm{MI}(r,n)>0\): masked prompting ranks gold tokens relatively higher (model benefits from mask-style probes) --- call these \emph{mask-friendly}.
		\item \(\mathrm{MI}(r,n)<0\): prompting (prefix LM) is relatively better; relation is \emph{mask-resistant}.
		\item The magnitude of MI indicates the strength of the relative advantage.
	\end{itemize}
	
	In low-resource practice we compute MI at small \(n\) (e.g., \(n=5\)), because differences are more informative for strategy selection in few-shot settings (see Figure \ref{fig:deep}). In our experiments we use a threshold at \(30\%\) (MI $ \geq $ 0.30 at \(n=5\)) to partition relations into two coarse groups:
	\[
	\begin{cases}
		\text{Mask-Filling} & \text{ if } \mathrm{MI}\ge 0.30,\\[3pt]
		\text{Map-Phrasal} & \text{ otherwise.}
	\end{cases}
	\]
	
		The naming of the two groups reflects their linguistic and modeling behavior. \emph{Mask-Filling} relations tend to produce short, lexically bounded completions (e.g., nominal or adjectival tails) that align well with denoising-style objectives and direct span recovery around a mask token. In contrast, \emph{Map-Phrasal} relations often express event-level or intentional mappings---for instance, clauses describing what a person ``wants,'' ``needs,'' or ``intends''---that require the model to generate longer, compositional phrases best captured through prefix-style prompting. 

\section{Experimental Setup}

Recent work such as UL2 \cite{ul2} has shown that denoising and autoregressive (causal) objectives represent complementary regimes of language modeling rather than disjoint paradigms. Motivated by this unified perspective, we treat both denoising and language modeling as compatible training modes and analyze their alignment using the Maskability Index (MI). Our main model, \texttt{T5-base}, was pretrained with a denoising objective and an auxiliary language modeling objective, followed by supervised fine-tuning on downstream reasoning templates. This makes it a suitable candidate for our evaluation.

We employ the AdaFactor optimizer with a fixed learning rate of 0.0001 throughout fine-tuning. The mini-batch size is set to 8 for \texttt{T5-base}, and training is performed for a total of 3 epochs.

We evaluate nine relations from ATOMIC2020 as a knowledge-base completion task: given a head and a relation, the model must generate or complete the corresponding tail. Template families and few-shot sample sizes follow the design described in preceding sections. DepthRank is computed on a held-out set of 100 heads per relation (each with up to three reference tails), and MI is computed at \(n=5\).

	\section{Results}
	
	\subsection{Maskability Index (MI) for selected ATOMIC2020 relations}
	Table~\ref{table:mi} shows MI computed at \(n=5\) for relations we analyze in this paper; the 30\% threshold separates Mask-Filling from Map-Phrasal groups.
	
	\begin{table}[ht]
		\centering
		\small
		\begin{tabular}{lcc}
			\toprule
			\textbf{Relation} & \textbf{MI (\%)} & \textbf{Assigned Group} \\
			\midrule
			AtLocation & 38.25 & Mask-Filling \\
			ObjectUse  & 35.47 & Mask-Filling \\
			xAttr      & 48.49 & Mask-Filling \\
						CapableOf      & 36.41 & Mask-Filling \\
									HasProperty      & 38.64 & Mask-Filling \\
												FilledBy      & 43.32 & Mask-Filling \\
			xIntent    & -65.74 & Map-Phrasal \\
			xNeed      & 3.92 & Map-Phrasal \\
			xWant      & 19.15 & Map-Phrasal \\
			\bottomrule
		\end{tabular}
		\caption{Maskability Index (MI) for selected ATOMIC2020 relations computed at \(n=5\). Relations with MI $\geq 30\%$ are grouped as ``Mask-Filling''.}
		\label{table:mi}
	\end{table}

	\begin{table}[t]
		\centering
		\small
		\begin{tabular}{l|cccccc|c}
			\toprule
			\textbf{Template / Task} & AtLoc & CapableOf & HasProp & ObjUse & FilledBy & xAttr & Avg. \\
			\midrule
			Masked Prompting (BERTScore) & 0.50 & 0.46 & 0.43 & 0.47 & 0.41 & 0.53 & 0.47 \\
			Masked Prompting (ROUGE)     & 0.20 & 0.20 & 0.18 & 0.20 & 0.12 & 0.18 & 0.18 \\
			Prompting (BERTScore) & 0.41 & 0.38 & 0.41 & 0.39 & 0.41 & 0.43 & 0.40 \\
			Prompting (ROUGE)     & 0.12 & 0.09 & 0.11 & 0.10 & 0.07 & 0.08 & 0.08 \\
			\bottomrule
		\end{tabular}
		\caption{Performance of T5-base on \textbf{mask-filling relations} (AtLoc, CapableOf, HasProp, ObjUse, FilledBy, xAttr).}
		\label{tab:mask_filling}
	\end{table}

	\paragraph{Qualitative observations}
Relations such as AtLocation, xAttr, and FilledBy are naturally mask-friendly: they typically involve short lexical heads and tails (e.g., locations, attributes, or functional fillers) that co-occur frequently in text and are well aligned with denoising-based pretraining objectives. In contrast, relations such as xIntent, xNeed, and xWant often require multi-step inference and longer phrasal realizations for both head and tail. For these relations, continuation-based prompting that supports longer semantic completions is more effective. Consistent with this distinction, MI correctly identifies these relations as less amenable to mask-style recovery in few-shot settings.

\begin{table}[t]
	\centering
	\small
	\begin{tabular}{l|ccc|c}
		\toprule
		\textbf{Template / Task} & xIntent & xNeed & xWant & All \\
		\midrule
		Masked Prompting (BERTScore) & 0.48 & 0.48 & 0.50 & 0.49 \\
		Masked Prompting (ROUGE)     & 0.40 & 0.34 & 0.35 & 0.36 \\
		Prompting (BERTScore) & \textbf{0.51} & \textbf{0.50} & 0.50 & 0.50 \\
		Prompting (ROUGE)     & \textbf{0.43} & 0.38 & 0.32 & 0.38 \\
		\bottomrule
	\end{tabular}
	\caption{Performance of T5-base on \textbf{map-phrasal relations} (xIntent, xNeed, xWant).}
	\label{tab:map_phrasal}
\end{table}
\section{Performance by MI category}

To better understand how the Maskability Index (MI) partitions relations according to their template affinity, we report generation quality of \texttt{T5-base} under both template families—Prompting (prefix) and Masked Prompting—using ROUGE and BERTScore as complementary metrics. These tables group relations according to their MI values (Table~\ref{table:mi}), separating \emph{mask-filling} and \emph{map-phrasal} types.

Both metrics exhibit a strong positive correlation ($r>0.9$ across all relations), confirming that they capture consistent aspects of semantic fidelity. The alignment between MI categories and empirical performance is evident: relations labeled as mask-filling achieve higher scores under masked prompting, while map-phrasal relations perform better under prefix prompting. This provides direct behavioral validation of MI as an indicator of template–objective alignment.

In Table~\ref{tab:mask_filling}, mask-filling relations such as \textit{AtLocation}, \textit{ObjectUse}, and \textit{xAttr} show a consistent performance advantage when evaluated with masked templates, particularly in ROUGE, suggesting that short lexical tails and high-frequency attribute words benefit from denoising-style recovery. Conversely, Table~\ref{tab:map_phrasal} shows that phrasal relations like \textit{xIntent}, \textit{xNeed}, and \textit{xWant} favor prefix prompting, aligning with their multi-token, inferential nature. These findings support our core hypothesis: MI not only predicts alignment but also explains variation in downstream generative success across template families.

Overall, the coherence between MI grouping and empirical metrics demonstrates that pretraining–prompt alignment manifests as measurable differences in generation quality. MI thus provides a practical bridge between intrinsic model statistics (DepthRank differences) and extrinsic evaluation outcomes (ROUGE, BERTScore), confirming its utility as a predictive diagnostic for template selection.

	\section{Discussion, limitations, and next steps}
	MI is a pragmatic, empirically grounded indicator for template or task–objective alignment. It is not a full theoretical guarantee — caveats include tokenization artifacts (very frequent tokens like ``to'' can bias DepthRank averages) and dependence on the base pretrained model. 
	
	Future work: (1) compute per-token or frequency-normalized MI variants, where each token’s contribution is weighted by its corpus frequency or normalized against its baseline predictability. (2) build an unsupervised estimator for MI from raw unlabeled corpora. At present, computing MI requires manually defined task templates and labeled examples, which limits its scalability and generality. An unsupervised variant could instead estimate task–objective compatibility intrinsically, by applying the model’s denoising or masking procedure to arbitrary text and aggregating token-level reconstruction likelihoods or DepthRank statistics. This would enable large-scale, task-agnostic diagnostics of model–objective alignment.

	\section{Related work} \label{sec:related}
	Probing PLMs as implicit knowledge bases began with cloze-style evaluations such as LAMA \cite{petroni2019language}, showing that factual and commonsense knowledge could be surfaced without fine-tuning. This line of work connected knowledge recoverability to pretraining objectives: masked/denoising models such as BERT and T5 \cite{devlin2019bert,raffel2020exploring} align with cloze templates, while autoregressive models such as GPT favor prefix-style continuation \cite{radford2019language,brown2020language,liu2021robust}. 
	While prior work has often compared masked and causal language modeling separately, UL2 \cite{ul2} introduced a hybrid objective that adaptively samples between them, demonstrating improved generalization across both generative and discriminative tasks. Our study complements this line of research by providing an empirical indicator that quantifies how well a given objective aligns with task structure.

A complementary direction enriches PLMs with \emph{external knowledge}. Retrieval-augmented methods integrate evidence sentences \cite{yu2022retrieval}, while commonsense KBs such as ConceptNet \cite{speer2016conceptnet}, ATOMIC \cite{sap2019atomic,hwang2020comet}, or extensions thereof \cite{choi2022kbc} provide structured resources. COMET \cite{bosselut2019comet} represents a hybrid strategy, dynamically generating KB-style facts from pretrained LMs; subsequent works \cite{ghazarian2023accent,tian2023harnessing} extend this to broader commonsense reasoning tasks.

	Building on these insights, \cite{zhang2021alleviating} introduced \emph{DepthRank}, demonstrating that triples vary in “depth” of commonsense abstraction and that deeper relations are harder to probe. Our work departs from depth alone by measuring template–objective alignment at the relation level.
	
	Parallel to probing studies, a rich literature has emerged on \emph{commonsense plausibility estimation} (CSPE). Early work showed that sentence probability and perplexity can serve as indicators of plausibility \cite{trinh2018simple,davison2019commonsense,liu2021commongen}. More recent approaches exploit LLMs’ instruction-following abilities to directly judge plausibility \cite{zhao2024llmcommonsense}, or evaluate hypotheses via entailment-path reasoning \cite{jung2022maieutic,tafjord2022entailer}. VERA \cite{liu2023vera} takes a discriminative approach, fine-tuning a classifier on millions of commonsense statements. By contrast, ComPaSS \cite{cui2025compass} operationalizes plausibility as a \emph{semantic shift} between anchor and candidate sentences, using similarity as a zero-shot plausibility score.
	
	Our proposed \emph{Maskability Index (MI)} complements these strands. Unlike CSPE methods that estimate the plausibility of individual statements, MI quantifies the relative advantage of masked versus prefix prompting for a given relation. This makes MI a diagnostic tool for template–objective alignment in generative knowledge completion, distinct from plausibility-based scoring or external knowledge augmentation.

	\section{Conclusion}
	We introduced the Maskability Index (MI), a DepthRank-based metric to predict whether a relation benefits from mask-style templates vs. prefix prompting in few-shot scenarios. MI helps automatically determine which adaptation strategy and template family to use. Future work will extend MI to per-token variants and build automatic MI estimators.


\begin{thebibliography}{10}
	
	\bibitem{bosselut2019comet}
	Antoine Bosselut, Hannah Rashkin, Maarten Sap, Chaitanya Malaviya, Asli
	Celikyilmaz, and Yejin Choi.
	\newblock Comet: Commonsense transformers for automatic knowledge graph
	construction.
	\newblock In {\em ACL}, 2019.
	
	\bibitem{brown2020language}
	Tom Brown, Benjamin Mann, Nick Ryder, Melanie Subbiah, Jared~D. Kaplan,
	Prafulla Dhariwal, Arvind Neelakantan, Pranav Shyam, Girish Sastry, Amanda
	Askell, et~al.
	\newblock Language models are few-shot learners.
	\newblock In {\em Advances in Neural Information Processing Systems (NeurIPS)},
	volume~33, pages 1877--1901, 2020.
	
	\bibitem{choi2022kbc}
	Byeongmin Choi, Yong-Sook Lee, Yeunwoong Kyung, and Eunchan Kim.
	\newblock Albert with knowledge graph encoder utilizing semantic similarity for
	commonsense question answering.
	\newblock {\em arXiv preprint arXiv:2211.07065}, 2022.
	
	\bibitem{cui2025compass}
	Wanqing Cui, Keping Bi, Jiafeng Guo, and Xueqi Cheng.
	\newblock Estimating commonsense plausibility through semantic shifts.
	\newblock {\em arXiv preprint arXiv:2502.13464}, 2025.
	
	\bibitem{davison2019commonsense}
	Joe Davison, Joshua Feldman, and Alexander~M. Rush.
	\newblock Commonsense knowledge mining from pretrained models.
	\newblock In {\em EMNLP-IJCNLP}, pages 1173--1178, 2019.
	
	\bibitem{devlin2019bert}
	Jacob Devlin, Ming-Wei Chang, Kenton Lee, and Kristina Toutanova.
	\newblock Bert: Pre-training of deep bidirectional transformers for language
	understanding.
	\newblock In {\em NAACL}, 2019.
	
	\bibitem{gao2021making}
	Tianyu Gao, Adam Fisch, and Danqi Chen.
	\newblock Making pre-trained language models better few-shot learners.
	\newblock In {\em Proceedings of the 59th Annual Meeting of the Association for
		Computational Linguistics (ACL)}, volume~1, pages 3816--3830. Association for
	Computational Linguistics, 2021.
	
	\bibitem{ghazarian2023accent}
	Sarik Ghazarian, Yijia Shao, Rujun Han, Aram Galstyan, and Nanyun Peng.
	\newblock Accent: An automatic event commonsense evaluation metric for
	open-domain dialogue systems.
	\newblock In {\em ACL}, 2023.
	
	\bibitem{atomic2020}
	Jena~D. Hwang, Chandra Bhagavatula, Ronan~Le Bras, Jeff Da, Keisuke Sakaguchi,
	Antoine Bosselut, and Yejin Choi.
	\newblock Comet-atomic 2020: On symbolic and neural commonsense knowledge
	graphs.
	\newblock In {\em Proceedings of the 35th AAAI Conference on Artificial
		Intelligence (AAAI)}, pages 6384--6392, 2021.
	
	\bibitem{hwang2020comet}
	Jena~D. Hwang, Chandra Bhagavatula, Ronan Le~Bras, Jeff Da, Keisuke Sakaguchi,
	Antoine Bosselut, and Yejin Choi.
	\newblock Comet-atomic 2020: On symbolic and neural commonsense knowledge
	graphs.
	\newblock In {\em AAAI}, 2020.
	
	\bibitem{jiang2020can}
	Zhengbao Jiang, Frank~F Xu, Jun Araki, and Graham Neubig.
	\newblock Can you tell me how to improve my prompt? learning to rephrase
	prompts for language models.
	\newblock In {\em Proceedings of the 58th Annual Meeting of the Association for
		Computational Linguistics}, pages 5925--5936. Association for Computational
	Linguistics, 2020.
	
	\bibitem{jung2022maieutic}
	Jaehun Jung, Lianhui Qin, Sean Welleck, Faeze Brahman, Chandra Bhagavatula,
	Ronan Le~Bras, and Yejin Choi.
	\newblock Maieutic prompting: Logically consistent reasoning with recursive
	explanations.
	\newblock In {\em EMNLP}, pages 1266--1279, 2022.
	
	\bibitem{liu2023vera}
	Jiacheng Liu, Wenya Wang, Dianzhuo Wang, Noah~A. Smith, Yejin Choi, and
	Hannaneh Hajishirzi.
	\newblock Vera: A general-purpose plausibility estimation model for commonsense
	statements.
	\newblock In {\em EMNLP}, pages 1264--1287, 2023.
	
	\bibitem{liu2021pretrain}
	Pengfei Liu, Weizhe Yuan, Jinlan Fu, Zhengbao Jiang, Hiroaki Hayashi, and
	Graham Neubig.
	\newblock Pre-train, prompt, and predict: A systematic survey of prompting
	methods in natural language processing.
	\newblock {\em arXiv preprint arXiv:2107.13586}, 2021.
	
	\bibitem{liu2021commongen}
	Yixian Liu, Liwen Zhang, Wenjuan Han, Yue Zhang, and Kewei Tu.
	\newblock Constrained text generation with global guidance: Case study on
	commongen.
	\newblock {\em arXiv preprint arXiv:2103.07170}, 2021.
	
	\bibitem{liu2021robust}
	Zhuang Liu, Wayne Lin, Ya~Shi, and Jun Zhao.
	\newblock A robustly optimized bert pre-training approach with post-training.
	\newblock In {\em China National Conference on Chinese Computational
		Linguistics}, pages 471--484. Springer, 2021.
	
	\bibitem{petroni2019language}
	Fabio Petroni, Tim Rockt{\"a}schel, Sebastian Riedel, Patrick Lewis, Anton
	Bakhtin, Yuxiang Wu, and Alexander Miller.
	\newblock Language models as knowledge bases?
	\newblock In {\em Proceedings of the 2019 Conference on Empirical Methods in
		Natural Language Processing}, pages 2463--2473, 2019.
	
	\bibitem{pouramini2024matching}
	Ahmad Pouramini and Hesham Faili.
	\newblock Matching tasks to objectives: Fine-tuning and prompt-tuning
	strategies for encoder-decoder pre-trained language models.
	\newblock {\em Applied Intelligence}, 54(20):9783--9810, Oct 2024.
	
	\bibitem{radford2019language}
	Alec Radford, Jeff Wu, Rewon Child, David Luan, Dario Amodei, and Ilya
	Sutskever.
	\newblock Language models are unsupervised multitask learners.
	\newblock Technical report, OpenAI, 2019.
	\newblock OpenAI Blog/Technical report.
	
	\bibitem{raffel2020exploring}
	Colin Raffel, Noam Shazeer, Adam Roberts, Katherine Lee, Sharan Narang, Michael
	Matena, Yanqi Zhou, Wei Li, and Peter~J. Liu.
	\newblock Exploring the limits of transfer learning with a unified text-to-text
	transformer.
	\newblock {\em Journal of Machine Learning Research}, 21(140):1--67, 2020.
	
	\bibitem{sap2019atomic}
	Maarten Sap, Ronan Le~Bras, Emily Allaway, Chandra Bhagavatula, Nicholas
	Lourie, Hannah Rashkin, Brendan Roof, Noah~A. Smith, and Yejin Choi.
	\newblock Atomic: An atlas of machine commonsense for if-then reasoning.
	\newblock In {\em AAAI}, pages 3027--3035, 2019.
	
	\bibitem{shin2020autoprompt}
	Taylor Shin, Yasaman Razeghi, Robert L~Logan IV, Eric Wallace, and Sameer
	Singh.
	\newblock Autoprompt: Eliciting knowledge from language models with
	automatically generated prompts.
	\newblock In {\em Proceedings of the 2020 Conference on Empirical Methods in
		Natural Language Processing (EMNLP)}, pages 4222--4235. Association for
	Computational Linguistics, 2020.
	
	\bibitem{speer2016conceptnet}
	Robyn Speer, Joshua Chin, and Catherine Havasi.
	\newblock Conceptnet 5.5: An open multilingual graph of general knowledge.
	\newblock In {\em AAAI}, 2016.
	
	\bibitem{tafjord2022entailer}
	Oyvind Tafjord, Bhavana Dalvi, and Peter Clark.
	\newblock Entailer: Answering questions with faithful and truthful chains of
	reasoning.
	\newblock In {\em EMNLP}, 2022.
	
	\bibitem{ul2}
	Yi~Tay, Mostafa Dehghani, Vinh~Q Tran, Xavier Garcia, Jason Wei, Xuezhi Wang,
	Hyung~Won Chung, Siamak Shakeri, Dara Bahri, Tal Schuster, et~al.
	\newblock Ul2: Unifying language learning paradigms.
	\newblock {\em arXiv preprint arXiv:2205.05131}, 2022.
	
	\bibitem{tian2023harnessing}
	Yufei Tian, Felix Zhang, and Nanyun Peng.
	\newblock Harnessing black-box control to boost commonsense in lms' generation.
	\newblock In {\em EMNLP}, 2023.
	
	\bibitem{trinh2018simple}
	Trieu~H. Trinh and Quoc~V. Le.
	\newblock A simple method for commonsense reasoning.
	\newblock {\em arXiv preprint arXiv:1806.02847}, 2018.
	
	\bibitem{yu2022retrieval}
	Wenhui Yu, Chenguang Zhu, Zhihan Zhang, Shuohang Wang, Zhuosheng Zhang, Yuwei
	Fang, and Meng Jiang.
	\newblock Retrieval augmentation for commonsense reasoning: A unified approach.
	\newblock In {\em EMNLP}, 2022.
	
	\bibitem{zhang2021alleviating}
	Yi~Zhang, Le~Li, Yuyang Wu, Qiang Su, and Xiao Sun.
	\newblock Alleviating the knowledge-language inconsistency: A study for deep
	commonsense knowledge.
	\newblock {\em arXiv preprint arXiv:2105.13607}, 2021.
	
	\bibitem{zhao2024llmcommonsense}
	Zirui Zhao, Wee~Sun Lee, and David Hsu.
	\newblock Large language models as commonsense knowledge for large-scale task
	planning.
	\newblock In {\em NeurIPS}, 2024.
	
\end{thebibliography}


\appendix

\section{Prompts for Knowledge Base Relations}
To convert knowledge-graph triples into natural language sentences, we used the relation-specific templates shown in Table~\ref{table:tuples}. Each template expresses a relation in a natural phrasing suitable for the \texttt{T5} model’s input format.

\begin{table*}[htbp]
	\centering
	\scriptsize
	\begin{tabular}{lll}
		\toprule
		\textbf{Relation} & \textbf{Natural Phrase} & \textbf{Example} \\
		\midrule
		AtLocation & located at & Book is located at the library \\
		ObjectUse & is used for & Hammer is used for building \\
		CapableOf & is capable of & Athlete is capable of running \\
		HasProperty & has the property of & The car has the property of being fast \\
		isFilledBy & is filled by & PersonX watches --- which is filled by the TV \\
		xAttr & is seen as & PersonX teaches at a university. PersonX is seen as intelligent \\     
		\midrule
		xIntent & because they intended & PersonX eats vegetables because they intended to be healthy \\
		xNeed & before that they need & PersonX attends the marathon; before that they need to train \\
		xWant & after that they want & PersonX washes the car; after that they want to dry it \\
		\bottomrule
	\end{tabular}
	\caption{Natural language templates used to verbalize different knowledge-base relations, along with illustrative examples.}
	\label{table:tuples}
\end{table*}

\end{document}